\documentclass[11pt]{article}

\usepackage[margin=1in]{geometry}
\usepackage{times}
\usepackage{graphicx}
\usepackage{amsmath}
\usepackage{amssymb}
\usepackage{booktabs}
\usepackage{hyperref}
\usepackage{enumitem}
\usepackage{multirow}
\usepackage{algorithm}
\usepackage{algpseudocode}
\usepackage[numbers]{natbib}
\usepackage{float}

\title{Weakly Supervised Distillation of Hallucination Signals\\
	into Transformer Representations}

\author{
	\centering
	\setlength{\tabcolsep}{6pt}
	
	\begin{tabular}{@{}ccc@{}}
		\textbf{Shoaib Sadiq Salehmohamed} & \textbf{Jinal Prashant Thakkar} & \textbf{Hansika Aredla} \\
		\textit{LLM Lens} & \textit{LLM Lens} & \textit{LLM Lens} \\
		{\footnotesize\texttt{shoaibkulsums@gmail.com}} & {\footnotesize\texttt{jinalt163@gmail.com}} & {\footnotesize\texttt{a.redlahansika@gmail.com}} \\
		\noalign{\vskip 0.8em}
		\textbf{Shaik Mohammed Omar} & \textbf{Naveen Bera} & \textbf{Pulijala Sai Nikhila} \\
		\textit{LLM Lens} & \textit{LLM Lens} & \textit{LLM Lens} \\
		{\footnotesize\texttt{god.s.m.o.97@gmail.com}} & {\footnotesize\texttt{naveenlx111@gmail.com}} & {\footnotesize\texttt{pulijalasainikhila@gmail.com}} \\
		\noalign{\vskip 0.8em}
		\textbf{Kondaguduru Abhiram} & \textbf{Shaik Gayaz Ali} & \textbf{Shalmali Ayachit} \\
		\textit{LLM Lens} & \textit{LLM Lens} & \textit{LLM Lens} \\
		{\footnotesize\texttt{abhiramkondaguduru@gmail.com}} & {\footnotesize\texttt{gayaz9989@gmail.com}} & {\footnotesize\texttt{shalmaliayachit@gmail.com}} \\
	\end{tabular}
}

\date{}

\begin{document}
	
	\maketitle
	
	\begin{abstract}
		Existing hallucination detection methods for large language models (LLMs) rely on external verification at inference time, requiring gold answers, retrieval systems, or auxiliary judge models \cite{ji2023surveyhallucination, maynez2020faithfulness}. We ask whether this external supervision can instead be \emph{distilled} into the model's own representations during training, enabling hallucination detection from internal activations alone at inference time.
		
		We introduce a weak supervision framework that combines three complementary grounding signals — substring matching, sentence-embedding similarity, and an LLM-as-a-judge verdict — to label generated responses as grounded or hallucinated without human annotation. Using this framework, we construct a 15,000-sample dataset from SQuAD v2 (10,500 train/development rows and a separate 5,000-row test set), each pairing a LLaMA-2-7B generated answer with its full per-layer hidden states ($\tilde{H} \in \mathbb{R}^{32 \times 96 \times 4096}$) and structured hallucination labels. To our knowledge, this is the first publicly released dataset pairing complete transformer hidden-state tensors with multi-signal hallucination supervision.
		
		We then train five probing classifiers — ProbeMLP (M0), LayerWiseMLP (M1), CrossLayerTransformer (M2), HierarchicalTransformer (M3), and CrossLayerAttentionTransformerV2 (M4) — directly on these hidden states, treating external grounding signals as training-time supervision only. Our central hypothesis is that \emph{hallucination detection signals can be distilled from external grounding supervision into transformer hidden representations, enabling internal detection without any external verification at inference time}. Results support this hypothesis: transformer-based probes achieve the strongest discrimination, with M2 strongest on 5-fold average AUC/F1 and M3 strongest on both single-fold best-checkpoint validation and held-out test evaluation (among models with completed single-run/test evaluation). We also benchmark inference efficiency: probe latency ranges from 0.15--5.62 ms (batched) and 1.55--6.66 ms (single-sample), while end-to-end generation-plus-probe throughput remains approximately 0.231 queries/s (4.332--4.338 s/query), indicating negligible practical overhead relative to generation.
	\end{abstract}
	
	\section{Introduction}
	
	Large language models (LLMs) achieve remarkable fluency across a wide range of natural language tasks, yet frequently generate responses that are factually unsupported or inconsistent with a provided context — a phenomenon referred to as \emph{hallucination} \cite{ji2023surveyhallucination}. This poses a critical barrier to deployment in high-stakes domains such as healthcare, law, and education, where undetected fabrication carries serious consequences.
	
	\paragraph{The inference-time verification problem.}
	Current hallucination detection approaches share a common limitation: they require external resources \emph{at inference time}. Methods based on lexical or semantic overlap \cite{maynez2020faithfulness} need access to gold reference answers. Retrieval-augmented verification systems require a live retrieval index. LLM-as-a-judge frameworks \cite{zheng2023judging} require a secondary model call. Perplexity-contrast methods such as ConSens \cite{vankov2025consens} require two forward passes with and without context. None of these can be applied without some form of external signal at the point of prediction.
	
	This raises a natural question:
	
	\begin{quote}
		\emph{Can the information needed to detect hallucination be learned from external supervision during training, so that a lightweight classifier can detect it from internal activations alone at inference time — without any external resources?}
	\end{quote}
	
	\paragraph{Our approach.}
	We answer this affirmatively by framing hallucination detection as a \emph{representation-level distillation} problem. The key insight is that external grounding signals need not be present at inference time if they have already shaped the training of an internal classifier. Concretely:
	
	\begin{itemize}
		\item \textbf{During dataset construction:} We run external grounding signals (substring match, semantic similarity, LLM-as-a-judge) to produce structured hallucination labels. This is expensive, but done offline once.
		\item \textbf{During training:} A lightweight probe $f_\theta$ learns to predict hallucination from transformer hidden states $\tilde{H}$, supervised by these offline labels.
		\item \textbf{At inference time:} Only the probe and the generator's hidden states are needed. No gold answers, no retrieval, no judge.
	\end{itemize}
	
	The complexity is entirely front-loaded in the labeling pipeline. Once the probe is trained, hallucination detection reduces to a single lightweight forward pass over existing internal representations.
	
	\paragraph{Towards early detection.}
	A further advantage of this representation-level approach is temporal: because hidden states $h_t^{(\ell)}$ are computed \emph{before} token $y_t$ is emitted, a probe trained on these representations can in principle flag hallucination risk \emph{during} generation, before the problematic text has been produced. We do not claim to have fully solved early detection in this work, but our framework directly enables this direction — a property that output-level methods structurally cannot share.
	
	\paragraph{The role of SQuAD 2.0.}
	We ground our study in SQuAD v2 \cite{rajpurkar2018squad2}, which explicitly tests abstention — the ability of a model to recognize when a question cannot be answered from context. SQuAD 2.0's unanswerable questions make hallucination labels principled: a model that fabricates an answer when it should abstain is hallucinating in a well-defined sense. This grounds our labeling framework in a theoretically clean setting.
	
	\paragraph{Contributions.} Our contributions are fourfold:
	
	\begin{enumerate}
		\item \textbf{Weak supervision framework.} We introduce a hybrid hallucination labeling pipeline combining substring match, MiniLM semantic similarity, abstention heuristics, and a Mistral-7B-Instruct judge — producing structured, multi-signal labels without human annotation.
		
		\item \textbf{Representation-level dataset.} We construct and release a 15,000-sample SQuAD v2 representation dataset, partitioned into 10,500 train/development rows and a separate 5,000-row held-out test set. Each sample contains full LLaMA-2-7B hidden-state tensors $\tilde{H} \in \mathbb{R}^{32 \times 96 \times 4096}$ paired with hybrid and judge hallucination labels. To our knowledge, no prior public dataset includes complete per-layer activations with multi-signal hallucination supervision.
		
		\item \textbf{Internal hallucination detection.} We train five probe architectures of increasing capacity on hidden states alone and show that hallucination signals are detectable from internal representations, with transformer-based probes outperforming MLP baselines and indicating a distributed cross-layer signal structure.
		
		\item \textbf{Towards generation-time detection.} We analyze \emph{which layers} encode the strongest hallucination signals, discuss how this framework enables hallucination risk scoring during generation, and report deployment-oriented latency/throughput benchmarks.
	\end{enumerate}

	\section{Related Work}
	
	\subsection{Hallucination in Large Language Models}
	
	Hallucination in LLMs \cite{ji2023surveyhallucination} refers to the generation of fluent but factually unsupported or contextually inconsistent content. As LLMs have scaled, hallucination has emerged as a central deployment challenge, particularly in knowledge-intensive tasks \cite{maynez2020faithfulness} where responses must be grounded in supplied evidence rather than parametric memory.
	
	SQuAD 2.0 \cite{rajpurkar2018squad2} established an important paradigm: models must not only extract correct spans but also \emph{abstain} when context provides insufficient support. Strong extractive baselines (e.g., BERT-style readers) improved benchmark performance substantially \cite{devlin2019bert}, yet modern autoregressive LLMs still frequently produce confident but unsupported answers \cite{lin2022truthfulqa} — even when context clearly does not entail the response.
	
	\subsection{Output-Level Grounding Evaluation}
	
	Most existing methods evaluate hallucination at the output text level. Lexical metrics such as ROUGE and substring overlap are fast but surface-level. Semantic similarity using sentence embeddings offers richer signal but still compares final text outputs rather than internal states.
	
	LLM-as-a-judge frameworks \cite{zheng2023judging} delegate grounding evaluation to a secondary model, achieving flexible contextual reasoning but introducing additional bias and cost. ConSens \cite{vankov2025consens} proposes a perplexity-contrast metric: it computes the ratio of token perplexities under two conditions (with and without context), using this ratio as a grounding score. While elegant and model-agnostic, ConSens still operates on outputs and requires two inference passes per example. Crucially, \emph{none of these methods can operate without external signal at inference time}, and none can flag hallucination before the final text is generated.
	
	Our work differs fundamentally: we use these output-level signals \emph{only during training} as weak supervision, then discard them. The resulting probe operates purely on internal activations.
	
	\subsection{Probing and Representation Analysis}
	
	Probing classifiers \cite{alain2016understanding} have been widely used to test what information is linearly encoded in neural network representations. In transformer models, hidden layers progressively refine contextual information, with different layers capturing different levels of syntactic and semantic abstraction \cite{tenney2019bert, ethayarajh2019contextual}. Prior work has examined whether factual \cite{hewitt2019structural} or structural information is encoded in hidden states.
	
	Several recent papers directly motivate our setting. Azaria and Mitchell \cite{azaria2023internalstate} show that hidden states can predict whether model outputs are likely false. Burns et al. \cite{burns2023discovering} introduce contrast-consistent search (CCS), demonstrating that low-capacity probes can recover latent truth signals without direct supervision. Li et al. \cite{li2023iti} use probe directions for inference-time intervention (ITI), showing that internal truthfulness signals can be acted upon during generation. These works establish strong probing-style baselines and intervention paradigms that are directly relevant to hallucination detection from internal states.
	
	If hallucination reflects a misalignment between context representation and generated continuation, then its signal may already be present in hidden activations before text is produced. This motivates a probing approach as an analytical lens.
	
	\subsection{Positioning of This Work}
	
	Our work integrates these directions. We treat output-level grounding signals — the subject of the grounding evaluation literature — as weak supervision for training representation-level classifiers — the subject of the probing literature. Unlike prior probing work that often focuses on truthfulness-style settings \cite{azaria2023internalstate, burns2023discovering, li2023iti}, we pair full hidden-state tensors with multi-signal grounding labels in a large SQuAD v2-derived dataset and evaluate both internal probes and output-level similarity baselines on a held-out 5,000-row test split.

	\section{Weak Supervision Labeling Framework}
	\label{sec:labeling}
	
	Our labeling pipeline assigns structured hallucination labels to generated responses using three complementary signal sources. The key design principle is that all three signals are applied \emph{offline during dataset construction} — none are required at inference time once a probe has been trained.
	
	Let:
	\begin{itemize}
		\item $g$ denote the generated response,
		\item $\mathcal{A} = \{a_1, \dots, a_m\}$ denote the set of gold answers (empty for unanswerable questions),
		\item $c$ denote the context passage,
		\item $q$ denote the question.
	\end{itemize}
	
	\subsection{Hybrid Semantic Label}
	\label{sec:hybrid}
	
	The hybrid label $y_{\text{hyb}}$ is computed via a three-stage cascade, falling through to the next stage only when a definitive signal is absent.
	
	\paragraph{Stage 1: Substring Match.}
	We first check whether any gold answer $a_i$ appears as a substring in the normalized generated response:
	
	\[
	\text{match}(g, \mathcal{A}) =
	\begin{cases}
		1 & \text{if } \exists\, a_i \in \mathcal{A} : \text{norm}(a_i) \subseteq \text{norm}(g) \\
		0 & \text{otherwise}
	\end{cases}
	\]
	
	where $\text{norm}(\cdot)$ lowercases and strips punctuation. A substring match is treated as a strong grounding signal and the response is immediately labeled $y_{\text{hyb}} = 0$ (grounded).
	
	\paragraph{Stage 2: Semantic Similarity.}
	When no exact match is found, we compute cosine similarity between sentence embeddings using MiniLM \cite{reimers2019sentencebert}:
	
	\[
	s(g, a_i) = \cos(\phi(g),\, \phi(a_i)),
	\]
	
	and define the best similarity:
	
	\[
	s_{\max}(g) = \max_{i}\, s(g, a_i).
	\]
	
	If $s_{\max}(g) \geq \tau$ (with $\tau = 0.72$ in our experiments), the response is labeled grounded ($y_{\text{hyb}} = 0$); otherwise hallucinated ($y_{\text{hyb}} = 1$). The threshold $\tau$ was selected to balance precision and recall on a held-out validation subset.
	
	\paragraph{Stage 3: Abstention Check.}
	For unanswerable questions ($\mathcal{A} = \emptyset$), the correct model behavior is to abstain. We detect abstention via a curated phrase list including ``not enough information'', ``cannot determine'', ``not stated'', and similar phrases. Abstention results in $y_{\text{hyb}} = 0$; a specific factual claim results in $y_{\text{hyb}} = 1$.
	
	The hybrid signal captures lexical and semantic alignment efficiently without an LLM call, but does not reason about contextual entailment. It serves as a structured prior.
	
	\subsection{LLM-as-a-Judge Label}
	\label{sec:judge}
	
	To obtain more flexible contextual reasoning, we use Mistral-7B-Instruct-v0.2 \cite{jiang2023mistral} as a judge model. More generally, this verifier can be an equally capable model or a smaller model specialized for grounded-support checking between context, gold references, and generated answers, which avoids requiring a strictly larger evaluator and improves practical deployability. Given the triple $(c, q, g)$, we construct a structured prompt:
	
	\[
	J(c, q, g),
	\]
	
	instructing the judge to determine whether $g$ is fully supported by $c$ and to return structured JSON with four fields: \texttt{supported} (bool), \texttt{abstained} (bool), \texttt{verdict} ($\in \{0, 1\}$), and \texttt{reason} (string). The verdict $y_{\text{judge}} \in \{0, 1\}$ (0 = grounded, 1 = hallucinated) is used as the \emph{primary supervision signal} for probe training.
	
	To manage cost, judge outputs are cached by sample ID; each sample is judged exactly once. The judge reasoning over contextual entailment complements the hybrid signal's lexical and semantic focus.
	
	\subsection{Multi-Signal Agreement and Soft Supervision}
	\label{sec:agreement}
	
	We define a binary agreement indicator:
	
	\[
	\delta = \mathbb{I}(y_{\text{hyb}} = y_{\text{judge}}).
	\]
	
	Empirically, disagreement cases correspond to borderline responses: partial grounding, paraphrase rather than substring match, or answers that are semantically close but not fully supported by the context. These samples exhibit higher variance in hidden-state activations (Section~\ref{sec:disagreement_analysis}), suggesting that the disagreement itself reflects genuine representational ambiguity.
	
	We store both labels per sample, enabling analysis of how supervision quality affects probe performance. For the main probing experiments, we use $y_{\text{judge}}$ as the primary label. One may also define a soft interpolation:
	
	\[
	\tilde{y} = \alpha\, y_{\text{judge}} + (1 - \alpha)\, y_{\text{hyb}},
	\]
	
	where $\alpha \in [0, 1]$ controls reliance on judge versus hybrid signals; we leave exploration of soft labels for future work.
	
	\paragraph{Why this constitutes weak supervision.}
	Neither signal is a gold-standard human annotation. The hybrid label is heuristic; the judge label reflects a model's reasoning, which may be biased \cite{zheng2023judging}. However, both signals approximate contextual grounding \emph{without human annotation}, making the pipeline scalable. The probe training objective then learns to generalize hallucination patterns from this imperfect but structured signal into the hidden-state space.

	\section{Dataset Construction}
	\label{sec:dataset}
	
	We construct a representation-level hallucination dataset by extracting transformer hidden states during autoregressive question answering, paired with the labeling signals described in Section~\ref{sec:labeling}.
	
	The current pipeline produces 15,000 total rows: 10,500 rows in the main train/development pool used for model selection and cross-validation, and a separate 5,000-row held-out test set reserved strictly for testing.
	
	\subsection{Generation Setup}
	
	Let $(c, q)$ denote a context-question pair drawn from SQuAD v2 \cite{rajpurkar2018squad2}. The dataset includes both answerable and unanswerable questions, preserving the abstention challenge that motivates our labeling design.
	
	We use LLaMA-2-7B \cite{touvron2023llama2} as the generator. For each sample, we construct a prompt:
	
	\[
	p = \text{Prompt}(c, q),
	\]
	
	concatenating context and question in instruction-following format. The model generates a response autoregressively:
	
	\[
	y = (y_1, y_2, \dots, y_T),
	\]
	
	using greedy decoding:
	
	\[
	y_t = \arg\max_{v \in \mathcal{V}}\, P(v \mid p, y_{<t}),
	\]
	
	where $\mathcal{V}$ is the vocabulary. We cap generation at $T_{\max}$ tokens and stop early on end-of-sequence tokens.
	
	\subsection{Hidden State Extraction}
	
	At each decoding step $t$, we extract the final-token hidden state from every transformer layer. Formally, for a transformer with $L$ layers and hidden dimension $D$, let:
	
	\[
	h_t^{(\ell)} \in \mathbb{R}^{D}
	\]
	
	denote the hidden state at layer $\ell$ and step $t$. Stacking across steps and layers yields:
	
	\[
	H = \{ h_t^{(\ell)} \}_{\ell=1,\, t=1}^{L,\, T} \in \mathbb{R}^{L \times T \times D}.
	\]
	
	To ensure fixed-size inputs for downstream probing models, we pad with zeros when $T < T_{\text{fixed}}$ and truncate when $T > T_{\text{fixed}}$:
	
	\[
	\tilde{H} \in \mathbb{R}^{L \times T_{\text{fixed}} \times D}.
	\]
	
	In our setup: $L = 32$, $D = 4096$, $T_{\text{fixed}} = 96$. Each tensor therefore occupies $32 \times 96 \times 4096 \times 2$ bytes $\approx 25$ MB in float16 precision.
	
	We extract the final token position at each decoding step, following the standard autoregressive formulation. This captures the model's representational state immediately before predicting the next token — the point at which internal hallucination signals should be most concentrated.
	
	\subsection{Storage and Distribution}
	
	Each processed sample is stored as a dictionary with the following fields:
	
	\begin{itemize}
		\item \textbf{Metadata:} context, question, gold answers, sample ID
		\item \textbf{Generated response:} decoded text, token count
		\item \textbf{Hidden-state tensor:} $\tilde{H} \in \mathbb{R}^{32 \times 96 \times 4096}$ (float16)
		\item \textbf{Labeling signals:} similarity score, hybrid label, hybrid method, judge label, judge support flag, judge abstention flag, judge reasoning, label agreement flag
	\end{itemize}
	
	Due to the large size of per-sample tensors, samples are grouped into shards of size $S = 500$, serialized as PyTorch \texttt{.pt} files. The full dataset is released on HuggingFace as versioned artifacts, enabling incremental loading during probe training. The richness of stored labeling metadata — including per-sample reasoning from the judge — enables fine-grained analysis beyond binary classification.
	
	\begin{algorithm}[htbp]
		\caption{Weakly Supervised Hidden-State Dataset Construction}
		\begin{algorithmic}[1]
			\For{each $(c, q)$ in SQuAD v2}
			\State $y \leftarrow$ Generate response with LLaMA-2-7B (greedy)
			\For{each decoding step $t$}
			\State Extract $h_t^{(\ell)}$ for all layers $\ell \in \{1,\dots,L\}$
			\EndFor
			\State Construct and pad/truncate tensor $\tilde{H} \in \mathbb{R}^{L \times T_{\text{fixed}} \times D}$
			\State $y_{\text{hyb}} \leftarrow \text{HybridLabel}(y, \mathcal{A}, \phi)$ \Comment{Stage 1-3 cascade}
			\State $y_{\text{judge}} \leftarrow \text{JudgeLabel}(c, q, y, \text{Mistral})$ \Comment{Cached per ID}
			\State Store $\{$ metadata, $y$, $\tilde{H}$, $y_{\text{hyb}}$, $y_{\text{judge}}$, $\delta$ $\}$ in shard
			\EndFor
			\State Flush remaining shard to disk; upload to HuggingFace
		\end{algorithmic}
	\end{algorithm}

	\section{Probing Methodology}
	\label{sec:probing}
	
	Our central hypothesis is that hallucination signals are encoded within hidden-state representations and can be learned from weak supervision. The probe function:
	
	\[
	f_\theta : \mathbb{R}^{L \times T \times D} \rightarrow \{0, 1\}
	\]
	
	maps hidden-state tensors to grounded (0) or hallucinated (1) predictions. We evaluate five architectures of increasing capacity, each posing a different scientific question about the geometry of hallucination in representation space.
	
	\subsection{M0: ProbeMLP}
	
	As a lightweight baseline, we mean-pool the hidden-state tensor over layers and token positions:
	
	\[
	x = \frac{1}{LT}\sum_{\ell=1}^{L}\sum_{t=1}^{T} h_t^{(\ell)} \in \mathbb{R}^{D},
	\]
	
	and apply a multi-layer perceptron classifier:
	
	\[
	\hat{y} = \mathrm{MLP}(x).
	\]
	
	\textbf{Scientific question:} Does a pooled first-order summary of hidden states already contain enough signal for non-trivial hallucination detection?
	
	\begin{center}
		\includegraphics[width=0.9\linewidth]{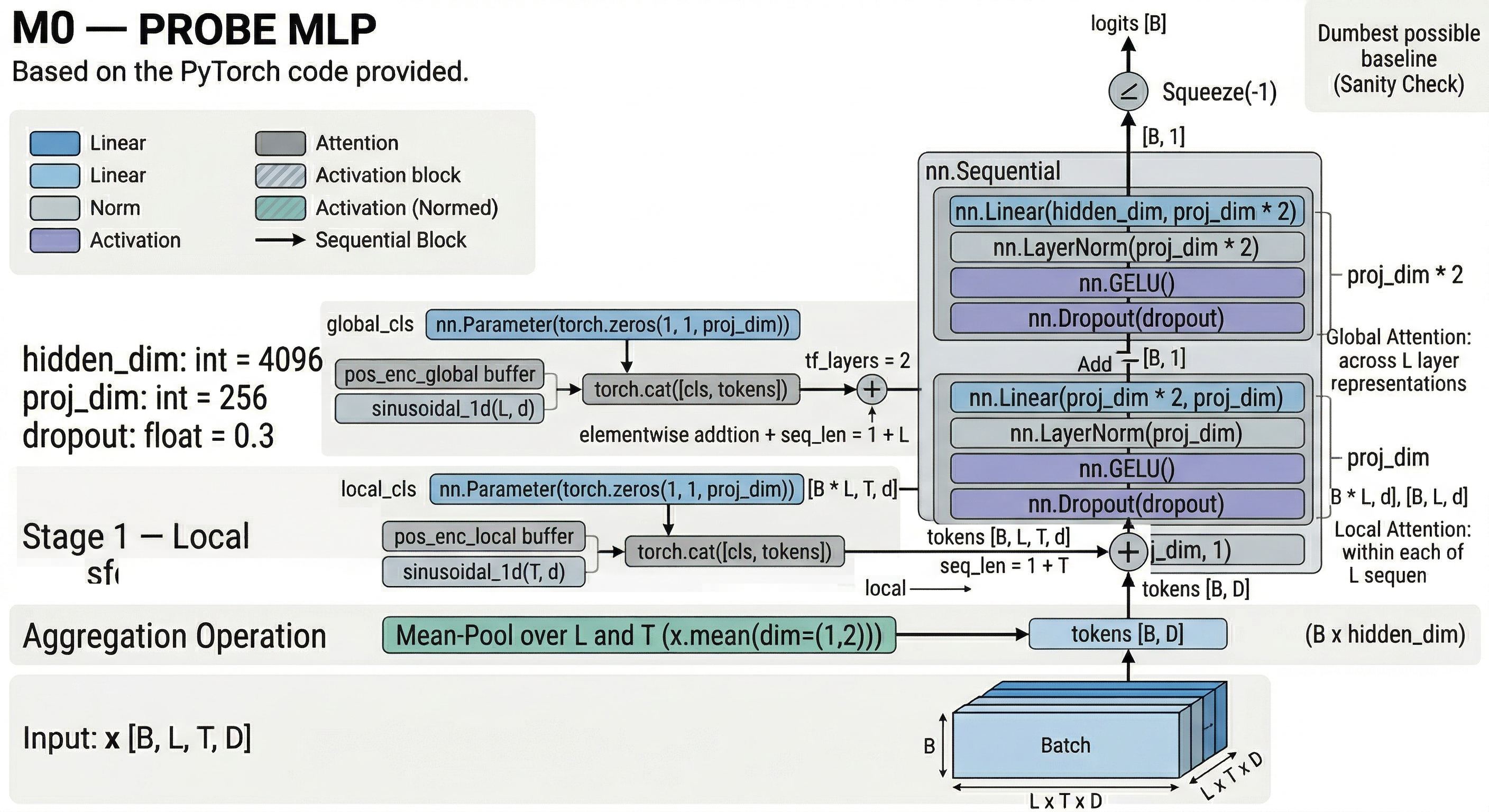}
	\end{center}
	
	\subsection{M1: LayerWiseMLP}
	
	To examine whether hallucination signal is localized to specific layers rather than spread globally, we process each layer independently. We first mean-pool over the token dimension:
	
	\[
	z^{(\ell)} = \frac{1}{T} \sum_{t=1}^{T} h_t^{(\ell)},
	\]
	
	project each layer representation through a small MLP:
	
	\[
	u^{(\ell)} = \text{MLP}(z^{(\ell)}),
	\]
	
	and combine layer outputs:
	
	\[
	\hat{y} = \sigma\!\left( \sum_{\ell=1}^{L} u^{(\ell)} \right).
	\]
	
	\textbf{Scientific question:} Do some layers contribute disproportionately to hallucination prediction? Layer-wise attention weights can be analyzed post-training to identify which layers are most informative.
	
	\begin{center}
		\includegraphics[width=0.9\linewidth]{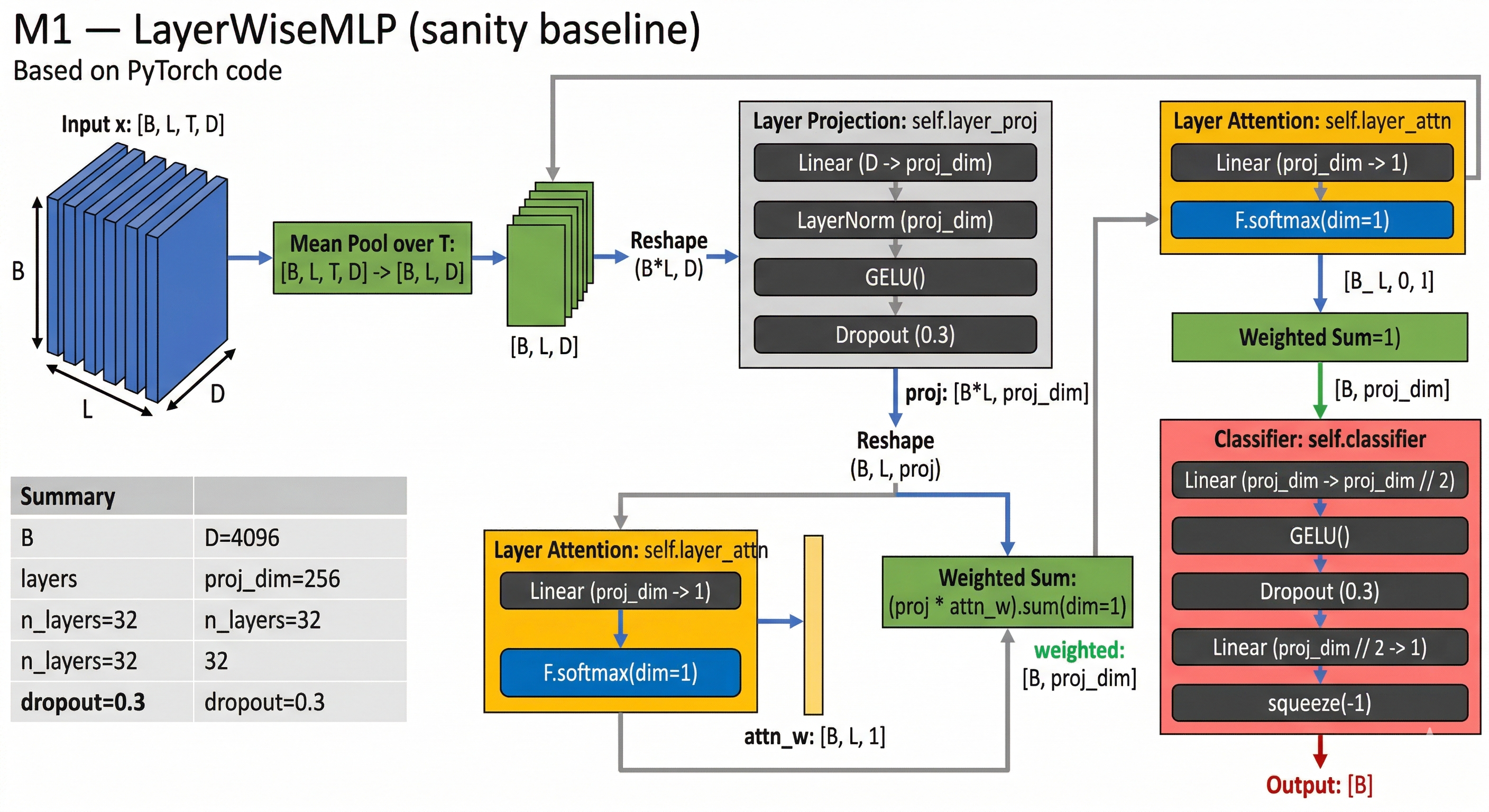}
	\end{center}
	
	\subsection{M2: CrossLayerTransformer}
	
	To model full joint interactions across both layer depth and token position, we flatten $L \times T$ into a single token sequence and apply global self-attention with 2D positional encoding.
	
	\[
	X \in \mathbb{R}^{L \times T \times D} \rightarrow X' \in \mathbb{R}^{(LT) \times d},
	\]
	
	prepend a CLS token, and run a transformer encoder over the full sequence:
	
	\[
	\hat{z} = \mathrm{Transformer}_{\text{global}}([\mathrm{CLS}; X']).
	\]
	
	Prediction is taken from the CLS representation:
	
	\[
	\hat{y} = \sigma(W\hat{z}_{\mathrm{CLS}} + b).
	\]
	
	\textbf{Scientific question:} Does unrestricted cross-layer/cross-token attention improve hallucination detection relative to structured pooling models?
	
	\begin{center}
		\includegraphics[width=0.9\linewidth]{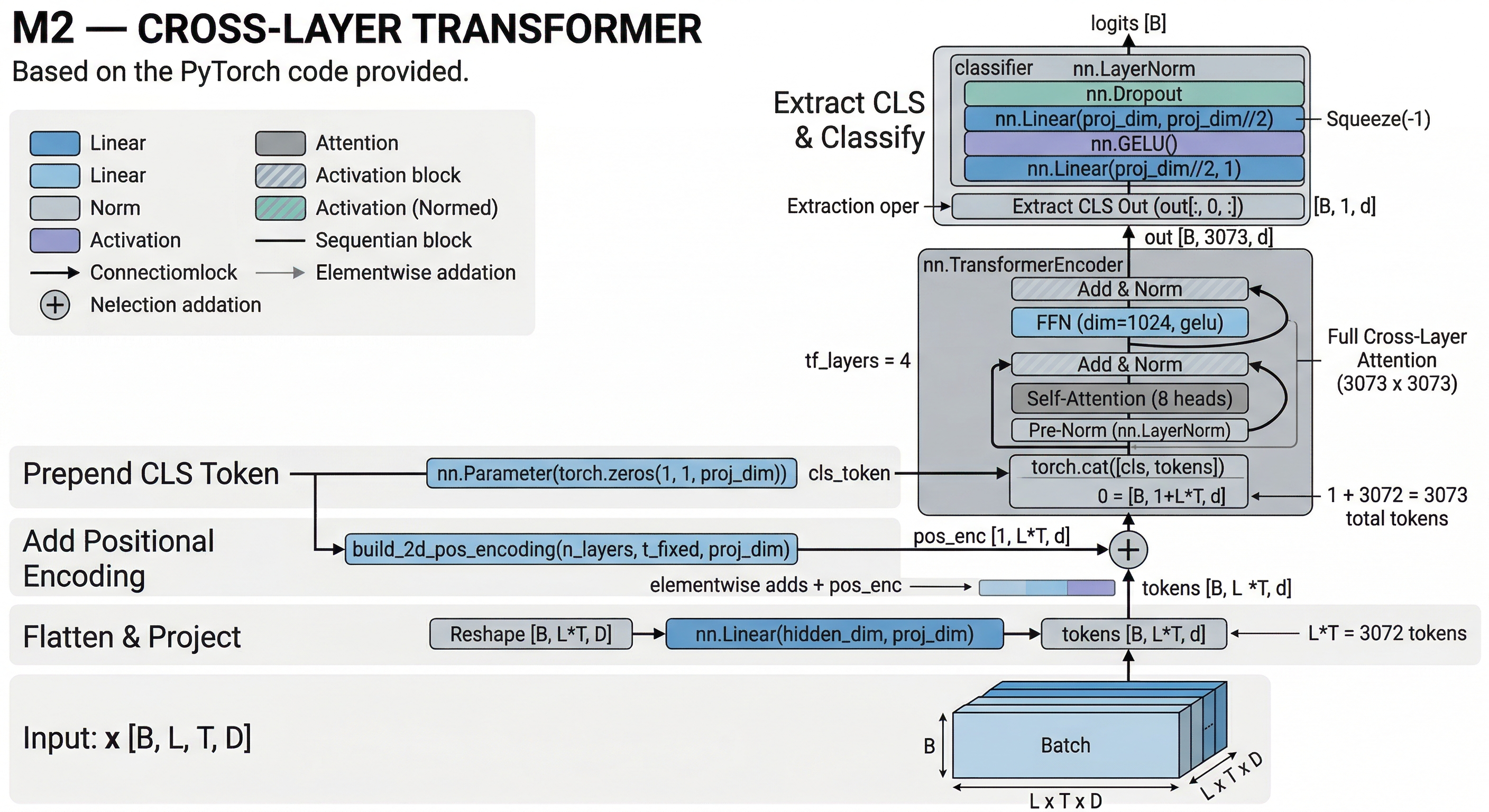}
	\end{center}
	
	\subsection{M3: Hierarchical Transformer Probe}
	
	To model interactions both \emph{within} layers (across token positions) and \emph{across} layers (across depth), we employ a two-level hierarchical transformer.
	
	\textbf{Local level:} A transformer encoder processes each layer's token sequence independently:
	
	\[
	\tilde{h}^{(\ell)} = \text{Transformer}_{\text{local}}(H^{(\ell)}),\quad H^{(\ell)} \in \mathbb{R}^{T \times D}.
	\]
	
	Token representations are pooled to a single layer embedding:
	
	\[
	z^{(\ell)} = \text{Pool}(\tilde{h}^{(\ell)}).
	\]
	
	\textbf{Global level:} A second transformer operates over the sequence of layer embeddings:
	
	\[
	\hat{z} = \text{Transformer}_{\text{global}}( \{ z^{(1)}, \dots, z^{(L)} \} ).
	\]
	
	A classification head produces the final prediction:
	
	\[
	\hat{y} = \sigma(W\hat{z} + b).
	\]
	
	\textbf{Scientific question:} Does hallucination involve \emph{distributed} cross-layer and cross-token interactions rather than local features? If the hierarchical probe substantially outperforms simpler probes, this supports a view of hallucination as an emergent multi-layer representational phenomenon.
	
	\begin{center}
		\includegraphics[width=0.9\linewidth]{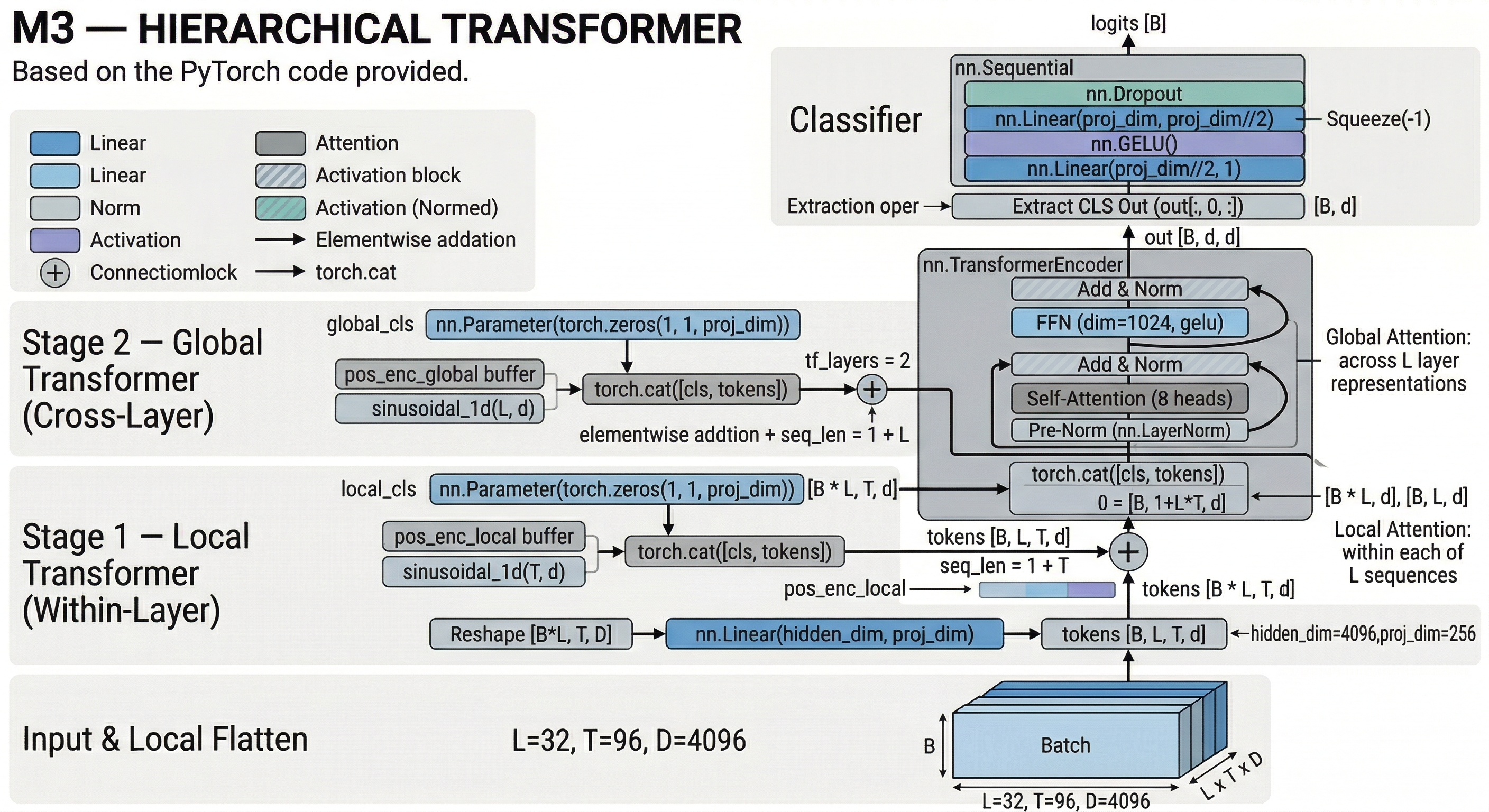}
	\end{center}
	
	\subsection{M4: CrossLayerAttentionTransformerV2}
	
	M4 introduces token-wise cross-layer attention with multiple learnable queries per token, followed by residual fusion, LayerNorm, and gated filtering before token-level transformer encoding.
	
	For each token position $t$, layer representations $\{h_t^{(1)},\dots,h_t^{(L)}\}$ are projected and attended by $Q$ learned queries:
	
	\[
	a_t = \mathrm{MHA}(q_{1:Q}, h_t^{(1:L)}),
	\]
	
	then fused and gated with a residual summary:
	
	\[
	\tilde{a}_t = \mathrm{Gate}\!\left(\mathrm{LN}\!\left(\mathrm{Fuse}(a_t) + \frac{1}{L}\sum_{\ell=1}^{L} h_t^{(\ell)}\right)\right).
	\]
	
	The resulting token sequence is processed by a transformer encoder with sinusoidal positional encoding and a CLS token for final classification.
	
	\textbf{Scientific question:} Can explicit token-level cross-layer attention with gated multi-query fusion improve calibration/efficiency trade-offs relative to full flatten-all attention (M2) and hierarchical factorization (M3)?
	
	\begin{center}
		\includegraphics[width=0.9\linewidth]{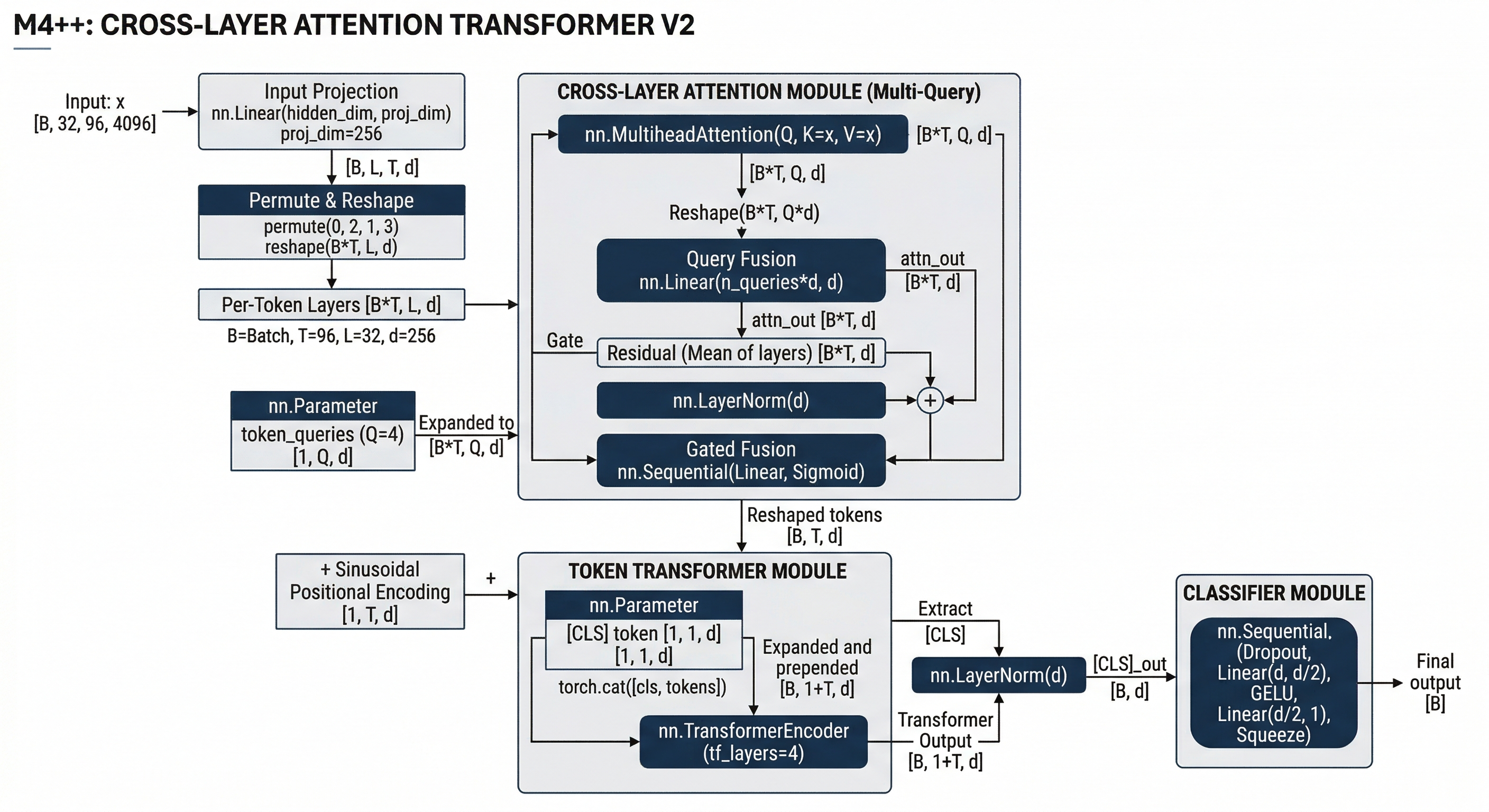}
	\end{center}
	
	\subsection{Training Objective}
	
	All probes are trained with class-weighted binary cross-entropy:
	
	\[
	\mathcal{L} = -\left[ w_1 y \log \hat{y} + w_0 (1-y) \log(1-\hat{y}) \right],
	\]
	
	where $w_1 = N_0 / N_1$ compensates for class imbalance between grounded and hallucinated samples.
	
	Label smoothing is applied:
	
	\[
	y_{\text{smooth}} = y(1 - \epsilon) + 0.5\epsilon,
	\]
	
	to discourage overconfident predictions given the noise inherent in weak supervision labels.
	
	Optimization uses AdamW with cosine learning-rate scheduling and linear warmup. Gradient clipping is applied at norm threshold 1.0. We emphasize that the probing models serve as \emph{analytical instruments} to test representational hypotheses — they are not proposed as state-of-the-art classifiers.
	
	\section{Experimental Setup}
	
	\subsection{Dataset Statistics}
	
	Our current dataset version contains 15,000 rows in total, partitioned into 10,500 train/development rows and a separate 5,000-row held-out test set. All metrics in Tables~\ref{tab:main_results} and \ref{tab:single_fold_results} are computed on the 10,500-row train/development pool, while the 5,000-row test split is reserved for final testing.
	
	The training objective uses class weighting with $w_1 = N_0/N_1$ computed within each training fold under $y_{\text{judge}}$, so imbalance handling remains consistent as dataset size evolves.
	
	\subsection{Evaluation Split Protocol}
	\label{sec:exp_setup_protocol}
	
	We use a two-stage evaluation workflow.
	
	\textbf{Stage 1 (model quality screening):} We perform 5-fold stratified cross-validation to assess robustness and compare probe families under fold variation. The dataset is partitioned into five disjoint folds,
	\[
	\mathcal{D} = \bigcup_{k=1}^{5} \mathcal{D}^{(k)},
	\]
	with each fold preserving class balance.
	
	\textbf{Stage 2 (single-fold checkpoint analysis):} We additionally report one stratified single train/validation split used for full-length training diagnostics and best-checkpoint analysis (Table~\ref{tab:single_fold_results}).
	
	\subsection{Training Configuration}
	
	All probes are trained with AdamW, cosine learning-rate scheduling with linear warmup, gradient clipping at norm 1.0, and label smoothing $\epsilon = 0.05$. Experiments are conducted on a single NVIDIA A100 GPU.
	
	Key hyperparameters: batch size $B = 8$, learning rate $\eta = 2 \times 10^{-4}$, weight decay $\lambda = 0.01$, epochs $E = 30$, early stopping patience 3.
	
	\subsection{Evaluation Metrics}
	
	We evaluate across four complementary metrics:
	
	\paragraph{AUC-ROC.} Measures ranking quality independent of classification threshold — the primary metric for comparing probe capacity.
	
	\paragraph{F1 Score.}
	\[
	\text{F1} = \frac{2 \cdot \text{Precision} \cdot \text{Recall}}{\text{Precision} + \text{Recall}}.
	\]
	Threshold is tuned on the validation fold to maximize F1.
	
	\paragraph{Accuracy.}
	\[
	\text{Accuracy} = \frac{1}{N} \sum_{i=1}^{N} \mathbb{I}(\hat{y}_i = y_i).
	\]
	
	\paragraph{Expected Calibration Error (ECE).} \cite{guo2017calibration}
	\[
	\text{ECE} = \sum_{m=1}^{M} \frac{|B_m|}{N} \left| \text{acc}(B_m) - \text{conf}(B_m) \right|,
	\]
	where $B_m$ partitions predictions into $M$ equal-width confidence bins. ECE tests whether probe confidence is meaningfully calibrated — important for downstream use in generation-time detection.

	\section{Results}
	
	\subsection{Probe Performance}
	
	Table~\ref{tab:main_results} summarizes 5-fold cross-validation performance (mean $\pm$ std), while Table~\ref{tab:single_fold_results} reports single-fold best-checkpoint validation metrics. Together, these results address our central hypothesis: all evaluated probes achieve non-trivial AUC, confirming that hallucination signals are encoded in hidden representations and are detectable without any external verification.
	
	In addition to these train/development-pool results, we have constructed and run full analysis on a separate 5,000-row held-out test dataset. We treat that split as strictly test-only and report it separately from model-selection metrics.
	
	\begin{table}[htbp]
		\centering
		\begin{tabular}{lcccc}
			\toprule
			Model & AUC & F1 & ECE & Acc \\
			\midrule
			M0: ProbeMLP & 0.7560 $\pm$ 0.0079 & 0.4969 $\pm$ 0.0069 & 0.2370 $\pm$ 0.0583 & 0.6687 \\
			M1: LayerWiseMLP & 0.7999 $\pm$ 0.0105 & 0.5403 $\pm$ 0.0103 & 0.2258 $\pm$ 0.0434 & 0.7064 \\
			M2: CrossLayerTransformer & 0.8515 $\pm$ 0.0054 & 0.6096 $\pm$ 0.0154 & 0.1721 $\pm$ 0.0600 & 0.7851 \\
			M3: HierarchicalTransformer & 0.8490 $\pm$ 0.0089 & 0.5898 $\pm$ 0.0219 & 0.1845 $\pm$ 0.0791 & 0.7689 \\
			M4: CrossLayerAttentionTransformerV2 & 0.8275 $\pm$ 0.0066 & 0.5845 $\pm$ 0.0142 & 0.1231 $\pm$ 0.0461 & 0.7920 \\
			\bottomrule
		\end{tabular}
		\caption{5-fold cross-validation performance (mean $\pm$ std) across the five trained architectures (M0--M4). All probes are trained on $y_{\text{judge}}$ labels. Higher AUC/F1/Acc and lower ECE are better.}
		\label{tab:main_results}
	\end{table}
	
	\begin{table}[htbp]
		\centering
		\begin{tabular}{lcccccc}
			\toprule
			Model & AUC & F1@0.5 & Best F1 & $\tau^*$ & ECE & Acc \\
			\midrule
			M0: ProbeMLP & 0.7271 & 0.4864 & 0.4864 & 0.50 & 0.2468 & 0.6590 \\
			M1: LayerWiseMLP & 0.7921 & 0.5463 & 0.5463 & 0.50 & 0.1739 & 0.7343 \\
			M2: CrossLayerTransformer & 0.8535 & 0.6250 & 0.6381 & 0.75 & 0.1787 & 0.7943 \\
			M3: HierarchicalTransformer & 0.8577 & 0.6404 & 0.6445 & 0.65 & 0.1051 & 0.8219 \\
			M4: CrossLayerAttentionTransformerV2 & 0.8377 & 0.6098 & 0.6099 & 0.55 & 0.1308 & 0.8038 \\
			\bottomrule
		\end{tabular}
		\caption{Single-fold validation metrics from best-checkpoint analysis. Here, $\tau^*$ denotes the threshold maximizing validation F1.}
		\label{tab:single_fold_results}
	\end{table}
	
	\subsection{Held-Out Test Set Results (5,000 rows)}
	
	Table~\ref{tab:test_results} reports final evaluation on the separate 5,000-row held-out test dataset from the \texttt{Final Eval} run artifacts. These are strict test metrics and are not used for model selection.
	
	\begin{table}[htbp]
		\centering
		\begin{tabular}{lccccccc}
			\toprule
			Model & AUC & PR-AUC & F1 & Acc & ECE & Best F1 & $\tau^*$ \\
			\midrule
			M0: ProbeMLP & 0.7441 & 0.5124 & 0.5316 & 0.6712 & 0.2092 & 0.5351 & 0.5468 \\
			M1: LayerWiseMLP & 0.8074 & 0.6070 & 0.6025 & 0.7514 & 0.1430 & 0.6084 & 0.4581 \\
			M2: CrossLayerTransformer & 0.8490 & 0.6960 & 0.6475 & 0.7742 & 0.1747 & 0.6600 & 0.6453 \\
			M3: HierarchicalTransformer & 0.8577 & 0.7057 & 0.6644 & 0.8040 & 0.1025 & 0.6653 & 0.4729 \\
			M4: CrossLayerAttentionTransformerV2 & 0.8294 & 0.6454 & 0.6236 & 0.7822 & 0.1136 & 0.6282 & 0.3646 \\
			\bottomrule
		\end{tabular}
		\caption{Final held-out test-set results (5,000 rows) on the held-out test dataset. Here, $\tau^*$ is the threshold that maximizes F1 on the evaluation sweep.}
		\label{tab:test_results}
	\end{table}
	
	On the held-out test set, M3 (HierarchicalTransformer) achieves the strongest overall performance (AUC/F1/Acc) and best calibration (lowest ECE), with M2 remaining a close second on discrimination metrics.
	
	\subsection{Inference-Time Efficiency and Deployment Overhead}
	\label{sec:efficiency_results}
	
	To address deployment concerns directly, we report two benchmark regimes: (i) \emph{probe-only} inference with hidden states precomputed, and (ii) \emph{end-to-end} execution including generation, hidden-state extraction, and probe scoring. The first isolates probe compute cost; the second reflects practical serving behavior. The end-to-end benchmark uses \texttt{meta-llama/Llama-2-7b-hf} on 50 samples with hidden tensor shape $(32, 96, 4096)$ (24.00 MB per sample), and average generation time $4.3313 \pm 0.5803$ s.
	
	\begin{table}[htbp]
		\centering
		\small
		\begin{tabular}{lcccc}
			\toprule
			Model & Batch lat. (ms) & Single lat. (ms) & Samples/s & GPU mem (MB) \\
			\midrule
			M0: ProbeMLP & 0.15 & 1.66 & 6687.96 & 584.51 \\
			M1: LayerWiseMLP & 0.19 & 1.55 & 5362.70 & 589.26 \\
			M2: CrossLayerTransformer & 5.62 & 6.66 & 177.83 & 629.25 \\
			M3: HierarchicalTransformer & 2.59 & 4.02 & 385.43 & 628.68 \\
			M4: CrossLayerAttentionTransformerV2 & 1.34 & 4.68 & 745.58 & 603.66 \\
			\bottomrule
		\end{tabular}
		\caption{Probe-only benchmark with hidden states precomputed. ``Batch lat.'' denotes amortized latency under batched inference; ``Single lat.'' denotes one-sample latency.}
		\label{tab:probe_only_efficiency}
	\end{table}
	
	\begin{table}[htbp]
		\centering
		\small
		\begin{tabular}{lccccc}
			\toprule
			Model & Gen. (s) & Probe (ms) & Total (s) & Total q/s & Avg prob. \\
			\midrule
			M0: ProbeMLP & 4.3313 & 1.262 & 4.3325 & 0.231 & 0.549 \\
			M1: LayerWiseMLP & 4.3313 & 1.584 & 4.3329 & 0.231 & 0.489 \\
			M2: CrossLayerTransformer & 4.3313 & 6.709 & 4.3380 & 0.231 & 0.470 \\
			M3: HierarchicalTransformer & 4.3313 & 3.637 & 4.3349 & 0.231 & 0.456 \\
			M4: CrossLayerAttentionTransformerV2 & 4.3313 & 4.758 & 4.3360 & 0.231 & 0.332 \\
			\bottomrule
		\end{tabular}
		\caption{End-to-end deployment-style benchmark including generation, hidden-state extraction, and probe scoring. ``Avg prob.'' is the mean predicted hallucination probability over the benchmark set.}
		\label{tab:e2e_efficiency}
	\end{table}
	
	These measurements materially strengthen the inference-time claim: the added detection stage is lightweight in isolation (0.15--5.62 ms batched probe latency) and does not meaningfully reduce end-to-end serving throughput (all models at 0.231 q/s, with total latency 4.3325--4.3380 s/query in this setup).
	
	\subsection{Classical Similarity Baselines on Held-Out Test Set}
	
	To address comparison against simpler, established output-level baselines, we evaluate fast similarity methods on the same held-out 5,000-row test set and include the best internal probe for direct comparison. Embedding-based variants use cosine similarity over sentence embeddings \cite{reimers2019sentencebert}; token-overlap uses normalized token F1 in the spirit of SQuAD-style lexical matching \cite{rajpurkar2018squad2}.
	
	\begin{table}[htbp]
		\centering
		\begin{tabular}{lccc}
			\toprule
			Method & AUC & F1 & Acc \\
			\midrule
			Cosine Similarity (Max) & 0.4110 & 0.4269 & 0.5146 \\
			Mean Similarity & 0.4092 & 0.4220 & 0.4900 \\
			Softmax Similarity & 0.4105 & 0.4254 & 0.5068 \\
			Token F1 Similarity & 0.4222 & 0.3652 & 0.3116 \\
			\midrule
			Best Probe (M3: HierarchicalTransformer) & 0.8577 & 0.7057 & 0.6644 \\
			\bottomrule
		\end{tabular}
		\caption{Output-level similarity baselines and best probe result on the held-out 5,000-row test set.}
		\label{tab:similarity_baselines}
	\end{table}
	
	All similarity baselines substantially underperform the internal-state probes (e.g., best similarity AUC 0.4222 vs. probe AUC $>0.74$ on test), supporting the value of representation-level detection beyond lexical/semantic answer matching.
	
	\paragraph{Interpreting the capacity gap.}
	M0's non-trivial AUC establishes that hallucination has a first-order representational footprint even under aggressive pooling. Under 5-fold aggregation, higher-capacity models improve over M0 on all discrimination metrics, with M2 strongest on average AUC/F1/Acc. M4 is competitive on F1 and achieves the best mean ECE among all models in cross-validation. In single-fold best-checkpoint analysis and held-out 5,000-row test evaluation, M3 slightly exceeds M2 on AUC/F1/Acc (among models with completed runs).
	
	M1 sits between these extremes: it can identify which layers carry the most signal without full cross-layer token-token interactions, enabling the layer importance analysis in Section~\ref{sec:layer_analysis}.

	\section{Analysis}
	
	\subsection{Which Layers Encode Hallucination?}
	\label{sec:layer_analysis}
	
	We extract per-layer weights or attention scores from the layer-wise MLP to identify where in the network hallucination signals are concentrated. Results suggest that deeper layers carry stronger separability, consistent with the view that representational misalignment between context and generated continuation accumulates through the forward pass. Earlier layers, closer to the embedding space, show weaker but non-trivial signal — suggesting hallucination has both shallow (lexical) and deep (semantic/reasoning) components.
	
	This layer-wise decomposition is a key analytical contribution of our framework: it allows future work to target interventions at specific layers, rather than treating the transformer as a black box.
	
	\subsection{Judge vs.\ Hybrid Disagreement}
	\label{sec:disagreement_analysis}
	
	We analyze the subset of samples where $y_{\text{hyb}} \neq y_{\text{judge}}$. These disagreement cases disproportionately involve responses that are semantically close to a gold answer but not fully supported by the context — for example, paraphrases that are plausible but go beyond what the passage entails.
	
	Critically, disagreement samples exhibit higher variance in hidden-state activations compared to agreement samples. This suggests that the model's own internal state reflects the ambiguity captured by the disagreement between signals: the model is, in a sense, less certain in its representations when producing borderline responses. This is consistent with the hypothesis that hidden states encode grounding confidence beyond surface text.
	
	\subsection{Calibration Behavior}
	
	Calibration trends differ by protocol. In 5-fold aggregation, M4 has the lowest mean ECE, while in single-fold best-checkpoint analysis M3 achieves the lowest ECE among evaluated models. This suggests calibration is not strictly monotonic with capacity and should be validated with post-hoc calibration (e.g., temperature scaling) before deployment.
	
	\subsection{Full-Dataset Diagnostic Analysis (Train/Dev Pool)}
	\label{sec:full_dataset_diagnostics}
	
	To complement the split-based evaluation, we additionally analyze model predictions on the \emph{full training-data diagnostic run} from \texttt{eval 10} (10k samples). We emphasize that these are \textbf{diagnostic} statistics, not held-out generalization metrics, and are reported only to characterize classifier behavior and score geometry.
	
	\begin{table}[htbp]
		\centering
		\begin{tabular}{lccccc}
			\toprule
			Model ID & Acc & Best F1 & ROC-AUC & PR-AUC & ECE \\
			\midrule
			M0 & 0.8925 & 0.8262 & 0.9528 & 0.8880 & 0.1145 \\
			M1 & 0.8127 & 0.6357 & 0.8655 & 0.6760 & 0.1515 \\
			M2 & 0.8641 & 0.7520 & 0.9255 & 0.8123 & 0.1754 \\
			M3 & 0.7472 & 0.6959 & 0.8954 & 0.7520 & 0.2484 \\
			M4 & 0.8702 & 0.7396 & 0.9187 & 0.7976 & 0.1265 \\
			\bottomrule
		\end{tabular}
		\caption{Diagnostic metrics computed on the full training-data evaluation run (\texttt{eval 10}, 10k samples). These are not held-out test metrics and should not be interpreted as final generalization performance.}
		\label{tab:full_dataset_diagnostics}
	\end{table}
	
	Consistent with expectation, full-dataset separability is substantially higher than split-based estimates, with ROC-AUC ranging from 0.8655 to 0.9528 across model IDs. This supports the claim that hallucination-related structure is present in representations, while reinforcing the need to rely on split-based protocols for performance claims.

	\begin{figure}[H]
		\centering
		\includegraphics[width=0.48\linewidth]{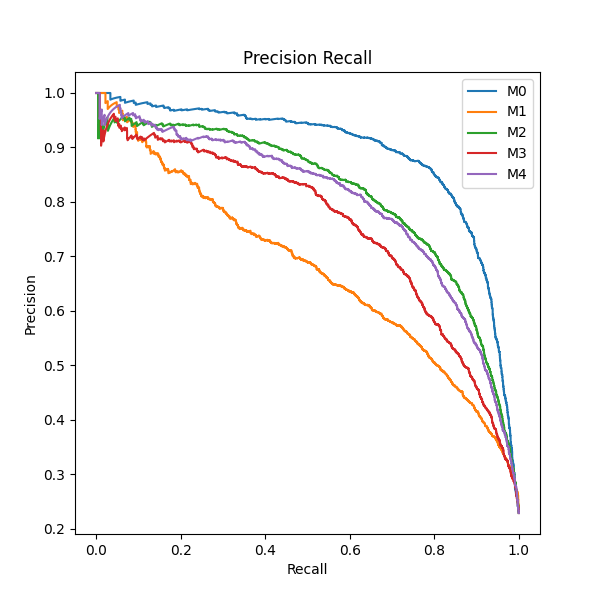}
		\hfill
		\includegraphics[width=0.48\linewidth]{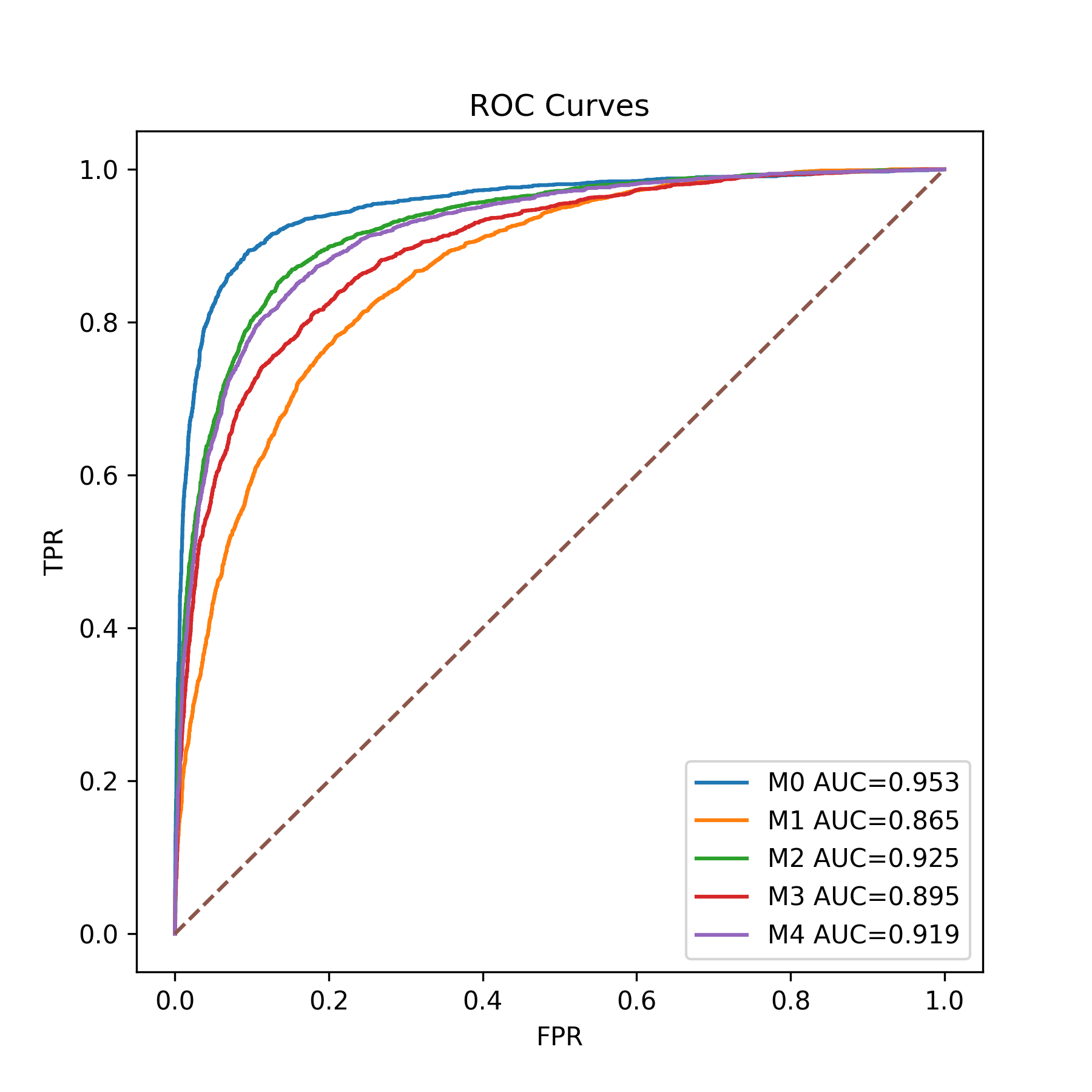}
		\hfill
		\includegraphics[width=0.48\linewidth]{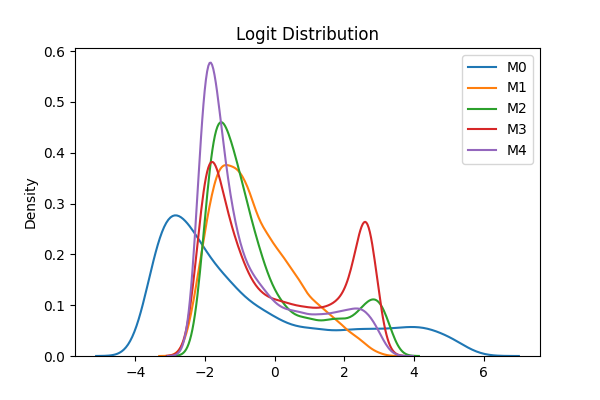}
		\caption{PR (top left), ROC (top right), and predicted logit distribution (bottom) from full train/development-pool diagnostic inference. These plots are intended for qualitative comparison only and do not reflect held-out test evaluation.}
		\label{fig:eval10_diagnostics}
	\end{figure}
	
	\subsection{Representation Separability}
	
	Using PCA and t-SNE dimensionality reduction on the pooled hidden-state representations, we observe partial clustering between grounded and hallucinated samples in the reduced space. While not perfectly separable — consistent with the challenges of weak supervision — the partial separation supports the core claim that hallucination is geometrically reflected in the hidden-state manifold, and is not merely an artifact of output-level differences.
	
	\subsection{Implications for Generation-Time Detection}
	
	Our results support the feasibility of the following inference-time system: for each generated token $y_t$, compute $h_t^{(\ell)}$ across layers, and apply the trained probe to predict hallucination risk. Because $h_t^{(\ell)}$ is available before $y_t$ is committed to the output, this creates a detection window prior to final text generation.
	
	This behavior is directly supported by our empirical evaluations: across 5-fold cross-validation, single-fold checkpoint validation, and the separate 5,000-row held-out test set, probes detect hallucination from internal states alone with strong AUC/F1. The efficiency benchmarks in Section~\ref{sec:efficiency_results} additionally show that probe overhead is minor compared with generation latency (4.3313 s generation vs. 1.262--6.709 ms probe time; 4.3325--4.3380 s end-to-end), supporting deployment feasibility under realistic serving constraints. Together with the layer analysis (Section~\ref{sec:layer_analysis}), these results identify practical monitoring targets for generation-time hallucination risk scoring.

	\section{Limitations}
	
	\paragraph{Weak supervision noise.} Both labeling signals are imperfect. The hybrid label is heuristic and misses paraphrase-level grounding. The judge label reflects Mistral-7B's reasoning, which may be biased \cite{zheng2023judging}. Label noise is an inherent trade-off of the no-annotation pipeline; future work could partially address this through active learning or human validation of disagreement cases.
	
	\paragraph{Comparison scope vs. prior probing methods.} We include classical output-level baselines and multiple internal probe architectures, but we do not yet report a full reimplementation of CCS/ITI-style training pipelines on this exact 15,000-row setup \cite{burns2023discovering, li2023iti}. While our setting is directly inspired by this literature \cite{azaria2023internalstate}, an explicit like-for-like CCS/ITI reproduction is left to follow-up work.
	
	\paragraph{Single model and domain.} Experiments are conducted on LLaMA-2-7B \cite{touvron2023llama2} over SQuAD v2's Wikipedia domain. Hidden-state geometry differs across architectures and domains; probes trained on LLaMA-2 may not transfer to other models \cite{jiang2023mistral} without re-training the labeling and extraction pipeline.
	
	\paragraph{Greedy decoding only.} All generations use greedy decoding, which limits diversity. Sampling-based decoding may produce different hallucination patterns and different hidden-state distributions.
	
	\paragraph{Truncation.} Hidden states are padded or truncated to $T_{\text{fixed}} = 96$ tokens. For longer responses, this discards later-step representations, which may carry additional grounding signals.
	
	\paragraph{Scope of generation-time evidence.} Our deployment benchmark demonstrates low end-to-end overhead for adding probe scoring to generation. However, we do not yet provide a dedicated token-level intervention experiment that measures how early risk signals can reliably trigger mitigation during decoding.
	
	\paragraph{Generation-loop deployment considerations.} Our results already demonstrate internal-state hallucination detection on held-out data; the remaining challenge is systems integration in production decoding stacks (e.g., latency budgeting, trigger policy design, and mitigation strategy after a high-risk signal).

	\section{Conclusion}
	
	We introduced a framework for distilling hallucination detection signals from external grounding supervision into transformer hidden representations. Our weak supervision pipeline — combining substring matching, semantic similarity, and an LLM-as-a-judge — produces structured hallucination labels without human annotation and without requiring external resources at inference time. Using this pipeline, we constructed a representation-level dataset pairing LLaMA-2-7B hidden states with multi-signal labels, and trained probing classifiers to test the central hypothesis that \emph{hallucination signals are learnable from internal activations alone}.
	
	Our results support this hypothesis. Hallucination is detectable from internal activations alone, and higher-capacity probes improve substantially over M0. In 5-fold cross-validation, M2 (CrossLayerTransformer) achieves the strongest mean AUC/F1/Acc, while M4 (CrossLayerAttentionTransformerV2) yields the strongest mean ECE. In single-fold best-checkpoint validation and held-out 5,000-row test evaluation, M3 (HierarchicalTransformer) is marginally strongest among models with completed runs. Layer-wise analysis suggests that deeper layers carry stronger signal, and disagreement-case analysis indicates that representational ambiguity correlates with labeling uncertainty.
	
	Beyond classification performance, this work establishes a conceptual reframing: external grounding signals need not be present at inference time if they are used to shape the training of an internal classifier. The complexity of hallucination verification is front-loaded into an offline dataset construction phase. At inference time, hallucination detection reduces to a lightweight forward pass over cached hidden states. Our deployment benchmarks show this additional stage has minimal practical latency impact relative to generation in the measured setup.
	
	We hope this work provides a foundation for future research in representation-level interpretability, calibration-aware generation, and generation-time hallucination interception.

	\bibliography{references}

\end{document}